# Diversification-Based Learning in Computing and Optimization

Fred Glover
Department of Electrical, Computer, and Energy Engineering
University of Colorado - Boulder
Boulder, CO, USA
glover@colorado.edu

Jin-Kao Hao [a,b]
a) LERIA, Université d'Angers, 2 bd Lavoisier, 49045 Angers, France
b) Institut Universitaire de France, 1 rue Descartes, 75231 Paris, France
jin-kao.hao@univ-angers.fr

## Abstract

Diversification-Based Learning (DBL) derives from a collection of principles and methods introduced in the field of metaheuristics that have broad applications in computing and optimization. We show that the DBL framework goes significantly beyond that of the more recent Opposition-based learning (OBL) framework introduced in Tizhoosh (2005), which has become the focus of numerous research initiatives in machine learning and metaheuristic optimization. We unify and extend earlier proposals in metaheuristic search (Glover, 1997, Glover and Laguna, 1997) to give a collection of approaches that are more flexible and comprehensive than OBL for creating intensification and diversification strategies in metaheuristic search. We also describe potential applications of DBL to various subfields of machine learning and optimization.

## 1. Introduction

Opposition-based learning (OBL) has become a source of numerous initiatives in the area of machine learning in artificial intelligence and associated initiatives to enhance metaheuristic search algorithms in optimization. Since its introduction in Tizhoosh (2005), a flood of proposals and studies have emerged to exploit its underlying ideas in a variety of contexts. (See for example, the surveys of Al-Qunaieer et al, 2010, Ergezer and Sikder, 2011, Xu et al., 2014a.)

An earlier framework introduced in the field of metaheuristic search (Glover, 1997; Glover and Laguna, 1997) provides a foundation that subsumes many of the OBL proposals, and gives a basis for additional enhancements. Starting from this foundation, we introduce a Diversification-Based Learning (DBL) framework that yields a collection of new strategies which enlarges those currently available in the OBL field. Accompanying this, we describe potential applications of DBL to various subfields of machine learning and optimization.

## 2. Background of Opposition-Based Learning

The notion of an "opposite number" or "opposite vector" in OBL bears a close relationship to the notion of a complemented solution in binary optimization. The original OBL definition is as follows.

**OBL Definition of an opposite number**

Relative to a given number $x' \in [L, U]$, the opposite number is given by $x'' = U + L - x'$.

It may be noted that in the case of a binary number $x' \in [0, 1]$, this definition corresponds precisely to the definition of the complement of $x'$ given by $x'' = 1 - x'$. The definition extends to the situation where $\mathbf{x'}$ is a vector, i.e., $\mathbf{x'} = (x_j': j \in N = \{1, \ldots, n\})$, by identifying bounds $L_j$ and $U_j$ for each component $x_j'$ of $\mathbf{x'}$, and generating a corresponding opposite value $x_j''$ for each $x_j'$ to give the components of an opposite vector $\mathbf{x''}$.

Subsequently, we will describe other definitions of an opposite number drawn from the OBL literature which we will compare to new definitions we propose that are motivated by metaheuristic considerations of diversification.

Historically, OBL has found applications in continuous optimization and has been used to reinforce a number of population evolutionary metaheuristics. This is typically achieved by coupling the generation of a candidate solution with the generation of its corresponding opposite solution during the population initialization and solution evolution phases. For instance, under the framework of differential evolution, OBL was employed to generate a diverse set of initial solutions and extend the current population by including their opposite solutions during the evolution process (Rahnamayan et al, 2008a). The same approach was also applied to other general methods like particle swarm optimization (Han and He, 2007), artificial neural networks (Ventresca and Tizhoosh, 2009), reinforcement learning (Tizhoosh, 2006), and population-based incremental learning (Ventresca and H.R.Tizhoosh, 2008). The idea of using OBL to solve discrete optimization problems has become an object of study in recent years. For instance, several authors have investigated OBL within the framework of Biogeography-based optimization to provide approximation methods for traveling salesman and graph coloring problems (Ergezer and D. Simon, 2011, Xu et al, 2014b). OBL was also combined with the memetic search framework to solve the maximum diversity problem (Zhou et al, 2017). In these studies, several alternative definitions of an opposite solution have been suggested to adapt the OBL concept to these specific problems.

## 3. Related Framework from Metaheuristic Search

As previously observed, the notion of an "opposite number" or "opposite vector" in OBL bears a close relationship to the notion of a complemented solution in binary optimization. As we will show in Section 4 below, there is a natural way to extend the definition of a binary complement to refer to numbers $x' \in [L, U]$ (for general lower and upper bounds L and U) that give a definition of an opposite number different from the OBL definition ($x'' = U + L - x'$) and that possesses useful features. First, however, we introduce a framework that can be used to exploit both the classical OBL definition and the new definitions we will subsequently introduce.

### 3.1 Opposite (Diverse) Collections Versus Opposite Solutions

Within the setting of binary optimization, the paper Glover (1997) proposes several diversification generators that start from an arbitrary binary vector $\mathbf{x}' = (x_j': j \in N = \{1, \ldots, n\})$ and create a *diverse collection* $D(\mathbf{x}')$ of additional vectors that differ from $\mathbf{x}'$ and from each other in various ways. (Each vector in $D(\mathbf{x}')$ is accompanied by its complement as a special case.) Consequently, from the perspective of opposition-based learning, this approach may be interpreted as replacing the notion of an opposite solution with the notion of a *diverse ("opposite") collection,* as embodied in the criteria for diversity used to create $D(\mathbf{x}')$. (The diversification generators in Glover (1997) for creating various collections $D(\mathbf{x}')$ are described in the Appendix, including generators to create diverse sets of permutation vectors in a sequencing context, to give a clearer idea of the kinds of criteria that can be relevant.)

### 3.2 Diverse Collections and Feasibility

A key insight for exploiting a collection of "opposites" embodied in a diverse collection comes from the observation that not all elements of a collection $D(\mathbf{x}')$ may be admissible or *feasible* relative to the requirements of a given setting – i.e., there may be constraints that exclude various element $\mathbf{x} \in D(\mathbf{x}')$ from being relevant.

Let $\mathbf{x}^o$ denote a solution drawn from $D(\mathbf{x}')\backslash\{\mathbf{x}'\}$ (which may or may not be the complement of $\mathbf{x}'$) and let $\mathbf{X}$ denote the set of feasible solutions. Then it becomes useful to create a mapping that transforms an infeasible vector $\mathbf{x}^o$ into a feasible vector which is "close to" $\mathbf{x}^o$.

For this, consider a *proximity function* $f^o(\mathbf{x})$ that embodies a measure of the proximity of $\mathbf{x}$ to $\mathbf{x}^o$. Then, for a given $\mathbf{x}^o \in D(\mathbf{x}')$ such that $\mathbf{x}^o$ is infeasible, we use a heuristic or exact method to

$$\text{Maximize } f^o(\mathbf{x}): \mathbf{x} \in \mathbf{X} \qquad (1)$$

The solution thus obtained will then take the place of $\mathbf{x}^o$ as a member of the diverse set $D(\mathbf{x}')$.

An example of $f^o(\mathbf{x})$ given in Glover (1997) for the binary case is the simple linear function

$$f^o(\mathbf{x}) = \sum(f_j^o x_j: j \in N) \qquad (2)$$

where $f_j^o > 0$ if $x_j^o = 1$ and $f_j^o < 0$ if $x_j^o = 0$. Thus, an optimal solution to (1), which maximizes $f^o(\mathbf{x})$ subject to $\mathbf{x} \in \mathbf{X}$, would set $x_j = 1$ for $f_j^o > 0$ (hence for $x_j^o = 1$) and set $x_j = 0$ for $f_j^o < 0$ (hence for $x_j^o = 0$) if such a solution were feasible, yielding $x^o$ itself.

For example, the simplest form of $f^o(\mathbf{x})$ is given by $f_j^o = 1$ if $x_j^o = 1$ and $f_j^o = -1$ if $x_j^o = 0$, thus producing the objective

$$\text{Maximize } \sum(x_j : j \in N: x_j^o = 1) - \sum(x_j : j \in N: x_j^o = 0)$$

By choosing positive and negative coefficients $f_j^o$ different from 1 and – 1 it becomes possible to produce solutions that possess various desirable features. In the context of metaheuristic optimization, for instance, it can be useful to allow these coefficients to embody intensification and diversification goals, as where a positive $f_j^o$ is made larger to more strongly emphasize setting $x_j = 1$ (or a negative $f_j^o$ is made smaller to more strongly emphasize setting $x_j = 0$) according to a frequency memory that counts the number of times $x_j = x_j^o$ in solutions of various categories (e.g., high quality solutions) found in the past. Such $f_j^o$ coefficients can be generated either deterministically or probabilistically as a function of frequency memory.

### 3.2.1 Useful and exploitable forms of X.

As an alternative to stipulating that **X** represents the set of feasible solutions to a particular problem, we can instead stipulate that **X** represents a set of solutions derived from a problem relaxation. In this case, a solution $\mathbf{x} \in \mathbf{X}$ that minimizes $f^o(\mathbf{x})$ can be taken as a starting point for metaheuristic or exact algorithms that generate fully feasible solutions. In the metaheuristic setting, such algorithms may be based on neighborhood search, where feasibility can be embodied in the definitions of the neighborhoods employed. In Section 5 we discuss the use of metaheuristics for generating such solutions in greater detail.

When $f^o(\mathbf{x})$ takes the form $\sum(f_j^o x_j: j \in N)$ indicated in (2), a number of commonly occurring types of constraints allow $f^o(\mathbf{x})$ to be optimized very simply. We identify a few examples as follows.

*Multiple Choice (GUB) Constraints*

These constraints are given by

$$\sum(x_j : j \in N_i) = 1, \ \ i \in M$$

where the sets $N_i, i \in M$ form a partition of N. Maximizing $f^o(x)$ over such constraints is accomplished by setting

$$x_{j(i)} = 1 \text{ for } j(i) \in N_i \text{ and } x_j = 0 \text{ for } j \in N_i \backslash \{j(i)\}$$

where $j(i) = \arg\max (f_j^o: j \in N_i)$. When $f_j^o$ has the elementary form where each $f_j^o$ is 1 or – 1, then any $j \in N_i$ with $f_j^o = 1$ qualifies as j(i), and otherwise every $j \in N_i$ qualifies as j(i) (since all coefficients $f_j^o$ for $j \in N_i$ are 0. Evidently, it is useful to differentiate among multiple optimal solutions by generating $f_j^o$ coefficients that differ from 1 and – 1 (again, for instance, determined probabilistically or deterministically according to values of $x_j$ in past solutions).

*Generalized Multiple Choice Constraints.*

A generalized instance of the foregoing GUB constraints takes the form

$$\sum(x_j : j \in N_i) = m_i, \ \ i \in M$$

where $0 < m_i < |N_i|$ and, as before, the sets $N_i$, $i \in M$, form a partition of N. The set of optimal solutions consist of those that satisfy

$$x_j = 1 \text{ if } j \in N_i(m_i),\ i \in M \text{ and } x_j = 0 \text{ otherwise}$$

where $N_i(m_i)$ consists of $m_i$ elements of $N_i$ having the largest values of $f_j^o$.

*A commonly encountered special case*

A frequently encountered version of the Generalized Multiple Choice Constraints occurs when M contains a single element, thus yielding a single constraint $\sum(x_j : j \in N_1) = m_1$. For the elementary instance of $f^o(\mathbf{x})$ where each $f_j^o$ is 1 or – 1, an optimal solution is constructed simply by observing that the set $N_1(m_1)$ is composed by selecting as many elements $j \in N$ as possible with $f_j^o = 1$ (equivalently, with $x_j^o = 1$) among the $m_1$ elements of $N_1$. The "opposite solution" proposed in Zhou et al. (2017) for the maximum diversity problem corresponds to such a solution.

*Additional useful and commonly occurring types of constraints*

Other kinds of constraints that often arise in practical settings, and that can easily be exploited by special cases of the preceding framework, include various types of network flow structures, especially those embodied in network assignment and distribution constraints. Optimal solutions in these instances can be obtained by standard network optimization algorithms. Likewise, a wide class of problems is attended by multiple knapsack constraints, and a variety of metaheuristic approaches can be used to obtain approximately optimal solutions to (1) in these situations.

## 4. An Alternative Definition of Opposite Solution from the DBL perspective

We first introduce a definition of an opposite solutions that generalizes the notion of a complementary solution and show that this definition has useful features that are missing from the classical OBL definition. Then we show how our definition can be embodied in a framework that generalizes the framework described in Section 3.

Once again consider the simplified situation where $x'$ is a number satisfying $x' \in [L, U]$. As an alternative to the OBL definition of an opposite value given by $x'' = U + L - x'$, we introduce the following notation.

Let $L^o$ and $U^o$ be values satisfying $L^o \geq L$ and $U^o \leq U$, with $L^o \leq U^o$, as where $L^o = L^o(\lambda_L) = L + \lambda_L(U - L)$ and $U^o = U^o(\lambda_U) = U - \lambda_U(U - L)$, for parameters $\lambda_L$ and $\lambda_U$ from the half-open interval [0, 0.5). (For example, if $\lambda_L = \lambda_U = 0.2$, then $L^o$ lies one fifth of the way from L to U and $U^o$ lies one fifth of the way from U to L.)[1]

[1] When $L^o$ and $U^o$ are given as functions of parameters $\lambda_L$ and $\lambda_U$, we normally do not choose $\lambda_L$ and $\lambda_U$ to be the same, which would cause $L^o$ and $U^o$ to differ by the same amount from L and U. The reason for this asymmetric treatment of $L^o$ and $U^o$ is because in many applications of optimization, L is given as a

Then we define the "opposite" point $x''$ associated with $x'$ to be the value of x in the interval $[L^o, U^o]$ that is farthest from $x'$. More precisely:

**DBL Opposite Definition**

$$x'' = L^o \text{ if } x' \geq (L^o + U^o)/2 \text{ and } x'' = U^o \text{ if } x' \leq (L^o + U^o)/2$$

We observe that whenever $x' > (L^o + U^o)/2$ the foregoing definition gives $x'' = L^o$ and whenever $x' < (L^o + U^o)/2$ the definition gives $x'' = U^o$. (These outcomes also hold if $x' > U^o$ and if $x' < L^o$.) Both $L^o$ and $U^o$ qualify as the opposite of x' when x' is the midpoint $(L^o + U^o)/2$, and in this case the tie between $L^o$ and $U^o$ can be settled arbitrarily.

To allow latitude in applying this definition, in the case where $L^o$ and $U^o$ are determined by reference to $\lambda_L$ and $\lambda_U$, these parameters can be varied for different components of a vector by choosing $\lambda_L$ and $\lambda_U$ randomly from chosen intervals (e.g., such as [1/6, 1/3] or [1/5, 2/5]). When $x'$ and $x''$ are required to be integers, we stipulate that $x''$ be assigned the integer value closest to the value indicated by the preceding DBL Opposite Definition. In the special case of binary vectors, this convention implies that different values of $\lambda_L$ and $\lambda_U$ from the half-open interval [0, 0.5) all are equivalent to defining $x''$ to be the complement of $x'$. However, different outcomes occur for more general continuous vectors and non-binary vectors.

The motivation for the preceding definition comes from two sources. First, this definition avoids a drawback of the classical OBL definition $x'' = U + L - x'$, as illustrated in the situation where $x' = 0.5(U + L)$. In this case the "opposite" of $x'$ is in fact the same as $x'$. (For example, when $x' \in [0, 1]$ and takes the midpoint value $x' = 0.5$, the opposite of $x'$ is also 0.5.) Moreover, the closer $x'$ is to the interval midpoint, the less that $x''$ differs from $x'$.

By contrast, according to our preceding definition, the values $x'' = L^o$ and $x'' = U^o$ both qualify as opposites of $x'$ when $x'$ lies halfway between $L^o$ and $U^o$, as previously noted. This holds for the special case where $L^o = L$ and $U^o = U$, which gives a direct comparison with the classical OBL definition.

More generally, we are motivated to choose $L^o$ and $U^o$ to differ from L and U because of an optimization strategy introduced in Glover and Martinson (1984) that progressively manipulates lower and upper bounds (therefore generating values that can be represented by $L^o$ and $U^o$) which enables a complex optimization problem to be solved by solving a series of much simpler problem relaxations.[2] Motivation is also provided by a parametric strategy for mixed integer programming (Glover, 2006) that imposes bounds through a parameterized objective function in

---

lower bound that is frequently attained (characteristically, L = 0), whereas U is typically chosen larger than any value that x will normally receive.

[2] In this case, the relaxations consisted of network relaxations.

place of a customary branching procedure, using adaptive memory strategies from tabu search to provide a control mechanism.[3]

Several alternative definitions of an opposite solution from the OBL literature that invite comparison with the DBL definition are offered in the papers Rahnamayan et al. (2008a), Ergezer et al. (2009), Ergezer and D. Simon (2011), Rahnamayan and Wang (2009) and Wang et al (2009). All but one of these definitions fail to drive x" away from x' in a manner that escapes being constrained by the midpoint of the interval, but instead generate a point at random that lies between this midpoint and another point (where the latter is either the point given by the classical OBL definition or the initial point x' itself). The definition that constitutes an exception, producing what is called the generalized opposite point in Wang et al. (2009), can be interpreted as attempting to drive x" away from the midpoint, but has the curious weakness of failing to be invariant under translations of the bounds L and U. In addition, this approach often generates a point x" that lies outside of the interval [L, U], which is then "repaired" by replacing x" with a point selected at random from the [L, U] interval.[4]

### Broadened Definitions and the Max-Min Distance Principle

We now consider several broader definitions of an opposite solution. In each of these, x" is chosen to be the point farthest from x' subject to being constrained to lie in a specified interval of values.

As in Section 3.1 above, we consider the relation of x" not just to a single value x' but to a diverse collection X', where we want x" to be in opposition to (diversified in relation to) all values x' in X'. A reason for introducing such a conception of opposition stems from the fact that in population-based metaheuristics we seek new solutions that are meaningfully opposed to all points in the population. Thus we return to the perspective where x' and x" are not single values, but vectors **x'** = ($x_1$', $x_2$', …, $x_n$') and **x"** = ($x_1$", $x_2$", …, $x_n$"). Then we apply the DBL definition of an opposite to the components $x_j$' and $x_j$" of these vectors. Thus, the set **X'** now represents a collection of vectors (such as a population or sub-population in a population-based metaheuristic) rather than a collection of values.

An approach suggested in Glover (1994) provides a starting point for this extension, in which the goal becomes to maximize the minimum distance of **x"** from all points **x'** $\in$ **X'**. A variation is to maximize a weighted sum of distances from the points **x'** $\in$ **X'**, but in this case the weights must be selected judiciously. A simple sum of distances can lead to generating vectors that have unattractive features from the standpoint of making **x"** meaningfully diverse relative to the points in **X'**.

Utilizing this perspective, we provide a simple component-by-component procedure for generating **x"** in opposition to (diverse from) the vectors in **X'**.

---

[3] Such strategies effectively augment diversification with intensification, and we later observe the relevance of joining these two processes in the present setting of a diversification-based approach for generating opposite solutions.

[4] One other definition, in Xu et al. (2011), exhibits complications similar to those of Wang et al. (2009).

**The Max-Min Principle**

For each component $x_j$ of **x**, write the corresponding values $x_j'$ of the vectors $\mathbf{x'} \in \mathbf{X'}$ as $x_j^1$, $x_j^2$, …, $x_j^r$, for $r = |\mathbf{X'}|$, where $L_j \le x_j^1 \le x_j^2, \ldots, \le x_j^r \le U_j$. For simplicity, define $x_j^0 = L_j$ and $x_j^{r+1} = U_j$.

**The Max-Min Opposite x" Relative to the set X'**

To determine each component $x_j''$ of **x"**, identify an index h, $0 < h \le r + 1$ that maximizes $x_j^h - x_j^{h-1}$. If h = 1, let $x_j'' = x_j^0$ and if h = r + 1, let $x_j'' = x_j^{r+1}$. Otherwise, let $x_j'' = (x_j^h + x_j^{h-1})/2$.

It is easy to verify that this determination of $x_j''$ maximizes the minimum distance from the values $x_j'$ for the vectors $\mathbf{x'} \in \mathbf{X'}$**.** Moreover, relative to each component $x_j'$ of **x'**, the result is equivalent to our earlier DBL definition if we stipulate that $x_j^0 = L_j^o$ and $x_j^{r+1} = U_j^o$, where $U_j^o$ receives a value that lies between $x_j^r$ and $U_j$ and $L_j^o$ receives a value that lies between $x_j^1$ and $L_j$.

In Section 6 we introduce definitions that apply to vectors as units rather than treating them component-by-component. Further use of the Max-Min Principle is described in the Appendix.

## 5. Generalizing the Framework of Section 3 to Handle Non-binary Vectors

As a starting point, we observe that we may apply the DBL definition of an opposite solution to map the binary solutions of a diverse collection produced by a diversification generator (such as one of those described in the Appendix) into a diverse collection applicable to vectors **x** where $x_j \in [L_j, U_j]$. Specifically, we operate as follows.

**Generating a diverse collection for non-binary vectors x**

Begin with an initial seed solution $\mathbf{x^s}$ and denote the collection of diverse solutions associated with $\mathbf{x^s}$ by $D^\#(\mathbf{x^s})$, where to begin $D^\#(\mathbf{x^s}) = \{\mathbf{x^s}\}$

Map $\mathbf{x^s}$ into a binary seed solution $\mathbf{y^s}$ where $y_j^s = 0$ if $x_j^s \le (L_j + U_j)/2$ and $y_j^s = 1$ if $x_j^s \ge (L_j + U_j)/2$

Apply a diversification generator to $\mathbf{y^s}$ to generate a diverse collection $D(\mathbf{y^s})$.

Map each solution $\mathbf{y^o} \in D(\mathbf{y^s})$ into a solution $\mathbf{x^o} \in D^\#(\mathbf{x^s})$ (i.e, add $\mathbf{x^o}$ to $D^\#(\mathbf{x^s})$) by one of the following rules:

(R1) Set $x_j^o = L_j^o$ if $y_j^o = 0$ and $x_j^o = U_j^o$ if $y_j^o = 1$.

(R2) Set $x_j^o = x_j^s$ if $y_j^o = y_j^s$ and otherwise set $x_j^o = L_j^o$ if $y_j^o = 0$ and $x_j^o = U_j^o$ if $y_j^o = 1$.

In the foregoing, it should be borne in mind that the values $L_j^o$ and $U_j^o$ of (R1) and (R2) may be chosen to take the form $L_j^o = L_j^o(\lambda_L) = L_j + \lambda_L(U_j - L_j)$ and $U_j^o = U_j^o(\lambda_U) = U_j - \lambda_U(U_j - L_j)$, as indicated in Section 4, where $\lambda_L$ and $\lambda_U$ are selected constants applied uniformly for all $j \in N$ or may be allowed to vary for each $j \in N$ (e.g., chosen randomly from a selected interval). We observe that if $\mathbf{x^s}$ is represented by $\mathbf{x'}$, then the opposite solution $\mathbf{x''}$ generated by the DBL definition is the same solution that will be generated by both (R1) and (R2) from the complement of $\mathbf{y^s}$.

**Generalizing the Feasibility Mapping of Section 3.**

In order to map a solution $\mathbf{x^o} \in D^{\#}(\mathbf{x^s})\backslash\{\mathbf{x^s}\}$ into a feasible solution, we generalize the approach of Section 3 as follows.

As in Section 3.2, we make use of a proximity function $f^o(\mathbf{x})$, which in this case we express in terms of a measure of distance between $\mathbf{x}$ and $\mathbf{x^o}$, and hence refer to minimization rather than maximization. Thus the objective becomes

$$\text{Minimize } f^o(\mathbf{x}): \mathbf{x} \in \mathbf{X} \qquad (3)$$

where, for example, $f^o(\mathbf{x})$ takes the form

$$f^o(\mathbf{x}) = \sum(f_j{}^o|x_j - x_j{}^o|): j \in N) \qquad (4)$$

and $f_j{}^o > 0$ for all $j \in N$. If $\mathbf{x^o}$ is binary, then (3) and (4) are equivalent to (1) and (2). By this means, we create diverse collections of solutions that satisfy $\mathbf{x} \in \mathbf{X}$. In the contexts normally used in OBL, where only a single opposite solution is generated for a given solution, it is natural to designate the opposite of $\mathbf{x^s}$ to be $\mathbf{x^o}$.

## 6. Uses of Metaheuristics to Generate Opposite Solutions

We identify several diversification methods for metaheuristics that are well-suited to generate solutions that qualify as opposite solutions. A useful feature of these methods is that the "opposite" solutions they produce retain feasibility when the metaheuristics yield feasible solutions (as where a neighborhood process preserves feasibility).

**Creating opposite solutions by neighborhood search with tabu search restrictions:** A diversification strategy from tabu search (see, e.g., Glover and Laguna, 1993, 1997) periodically introduces a large tabu tenure to prevent the trajectory of neighboring solutions from reversing any of its component moves. By selecting the number of moves defined by "large," it is possible to generate solutions that constitute varying levels of opposition relative to the solution that launched the trajectory. An extreme version of such an approach that uses an unbounded tabu tenure, and continues until no more moves are available to be selected, was found to be highly effective in Kelly et al. (1994). This outcome suggests that the use of large tabu tenures to identify opposite solutions deserves further exploration.

**Bi-directional opposite solutions from exterior path relinking:** Exterior path relinking (Glover, 2014; Duarte et al. 2015), is a population-based approach utilizing an initiating solution $\mathbf{x^I}$ and a guiding solution $\mathbf{x^G}$, to creates a trajectory from $\mathbf{x^I}$ to $\mathbf{x^G}$ that goes beyond the guiding solution $\mathbf{x^G}$. By interchanging the roles of the initiating and guiding solutions, the process may be viewed as creating an opposite solution from the pairing $(\mathbf{x^I}, \mathbf{x^G})$ in one direction and also from the pairing $(\mathbf{x^G}, \mathbf{x^I})$ in the reverse direction, to create a bi-directional determination of opposite solutions. The path relinking approach can also be applied with multiple guiding solutions, and can be varied by choosing different distances beyond $\mathbf{x^G}$ (or $\mathbf{x^I}$) for generating the opposite solution. These distances have a built-in limit which identifies an "extreme opposite" analogous to complementing a 0-1 vector.

**Generating opposite solutions from clustering**: Clustering provides an important opportunity for organizing the generation of opposite solutions by metaheuristic processes (see, e.g., Glover, 1977; Glover and Laguna, 1993). For example, exterior path relinking trajectories can be created that select initiating and guiding solutions from different clusters to induce a stronger diversification effect than choosing them from a common cluster. Moreover, when initiating solutions and guiding solutions are generated this way, a solution generated on an interior trajectory (that is, a solution between $\mathbf{x}^{\mathbf{I}}$ and $\mathbf{x}^{\mathbf{G}}$) that lies outside of the clusters can also qualify as being in opposition to $\mathbf{x}^{I}$ and $\mathbf{x}^{\mathbf{G}}$, according to its distance to the boundaries of the clusters containing $\mathbf{x}^{\mathbf{I}}$ and $\mathbf{x}^{\mathbf{G}}$.[5] This provides an additional distinction for classifying opposite solutions according to their intensification/diversification focus and invites research into the effect of this classification on generating useful opposite solutions in various contexts.

**Creating opposite solutions by extracting diverse subsets from larger populations:**
An alternative approach that provides a further basis for creating opposite solutions arises by generating a relatively large population by initial diversification strategies and then extracting mutually diverse subsets of points, based on criteria such as maximizing the minimum distance from other points in the set under construction. In a sequential procedure for extracting the points (Glover, 1994), these criteria can produce many ties for the element to be selected next, once a small number of elements have been selected. Even in the absence of ties such a constructive approach can ultimately create collections of points that can be improved according to the diversity criteria employed.

Drawing on this observation, a variety of more sophisticated iterative approaches are introduced in Glover (2016) for obtaining collections of points that are more diverse than those found by simpler constructive methods. These approaches, which are developed in the context of creating seed points for clustering, can equally be applied in other contexts to produce solutions that satisfy useful criteria of opposition.

## 7. Conclusions

The notion of "opposition" that gives rise to the definitions of an opposite solution introduced in opposition based learning (OBL) can be significantly extended by reference to earlier notions of diversification that have emerged in the area of metaheuristic search. The resulting diversification based learning (DBL) framework is not only more flexible than OBL, but overcomes limitations in the OBL definitions of an opposite solution. The DBL perspective further broadens the notion of opposition by conceiving it to refer not only to a single solution as an opposing partner of a given solution, but to refer to an "opposite collection" of solutions, as obtained by a diversification generator. These alternative notions lead to a model that allows the concept of opposition to operate within the context of feasibility, which is missing from the OBL framework except by the device of simply rejecting an infeasible opposite solution as inadmissible, without offering a direct means of establishing a connection to feasibility. Finally, we demonstrate how earlier diversification ideas for binary vectors can be generalized in the DBL framework to apply equally to non-binary vectors, identifying a range of metaheuristic procedures that can produce solutions that can meaningfully qualify as opposite solutions. The enhanced scope and adaptability of DBL opens the possibility of creating applications of this framework in realms where OBL has been too narrow to find a use, and invites studies in the area of metaheuristics where the principles underlying DBL remain largely unexplored.

[5] The distance from a point x to a cluster boundary in this case can be defined as the distance to the point in the cluster closest to x.

**References**

F. S. Al-Qunaieer, H. R. Tizhoosh and S. Rahnamayan (2010) "Opposition based computing - A survey," in: *Proceedings of International Joint Conference on Neural Networks (IJCNN-2010),* pp. 1–7.

A. Duarte, J. Sánchez-Oro, M.G.C. Resende, F. Glover and R. Marti (2015) "Greedy randomized search procedure with exterior path relinking for differential dispersion minimization," *Information Sciences,* Vol 296, pp. 46-60.

M. Ergezer and I. Sikder (2011) "Survey of oppositional algorithms," in: *Proceedings of International Conference on Computer and Information Technology*, 22–24 December, Dhaka, Bangladesh, pp. 623-628.

M. Ergezer, D. Simon and D. W. Du (2009) "Oppositional biogeography-based optimization," in: Proceedings of *IEEE International Conference on Systems, Man and Cybernetics*, San Antonio, USA, pp. 1009-1014.

M. Ergezer and D. Simon (2011) "Oppositional biogeography-based optimization for combinatorial problems," in: *Proceedings of Congress on Evolutionary Computation (CEC-2011),* pp. 1496–1503.

F. Glover (1977) "Heuristics for Integer Programming Using Surrogate Constraints," *Decision Sciences*, Vol. 8, No. 1, pp. 156-166.

F. Glover (1994) "Tabu Search for Nonlinear and Parametric Optimization (with Links to Genetic Algorithms)," *Discrete Applied Mathematics,* Vol 49, pp. 231-255.

F. Glover (1997) "A Template for Scatter Search and Path Relinking," in *Artificial Evolution, Lecture Notes in Computer Science,* 1363, J.-K. Hao, E. Lutton, E. Ronald , M. Schoenauer and D. Snyers, Eds. Springer, pp. 13-54.

F. Glover (2006) "Parametric Tabu Search for Mixed Integer Programs," *Computers and Operations Research,* Vol 33, Issue 9, pp. 2449-2494.

F. Glover (2014) "Exterior Path Relinking for Zero-One Optimization," *International Journal of Applied Metaheuristic Computing*, Vol. 5, Issue 3, pp. 1-8.

F. Glover (2016) “Pseudo-Centroid Clustering,” *Soft Computing*, ( ), pp. 1-22, Springer, published online 13 October 2016, content sharing version: http://rdcu.be/k1kY. DOI 10.1007/s00500-016-2369-6

F. Glover (2017) “Diversification Methods in 0-1 Optimization,” *Cornell University Library, arXiv: 1701.08709 [cs.AI]*, January 2017.

F. Glover and M. Laguna (1993) "Tabu Search," chapter in *Modern Heuristic Techniques for Combinatorial Problems*, C. Reeves, ed., Blackwell Scientific Publishing, pp. 71-140.

F. Glover and M. Laguna (1997) *Tabu Search*, Kluwer Academic Publishers, Springer.

F. Glover, R. Glover and F. Martinson (1984) “A netform system for resource planning in the U.S. bureau of land management,” *Journal of the Operational Research Society*, Vol. 35, No. 7, pp. 605-616.

L. Han and X.S. He (2007) “A novel opposition-based particle swarm optimization for noisy problems,” in *Proceedings of International Conference on Natural Computation*, 24–27 August, Haikou, China, pp. 624–629.

J.P. Kelly, M. Laguna and F. Glover (1994) "A study of diversification strategies for the Quadratic Assignment Problem," *Computers and Operations Research*, Vol. 21, No. 8, pp. 885-893.

S. Rahnamayan, H. R. Tizhoosh and M. Salama (2008a) “Opposition-based differential evolution,” *IEEE Transactions Evolutionary Computation*, Vol. 12, No. 1, pp. 64–79.

S. Rahnamayan, H. R. Tizhoosh and M. M. Salama (2008b) “Opposition versus randomness in soft computing techniques,” *Applied Soft Computing*, Vol. 8, No. 2, pp. 906–918.

S. Rahnamayan and G. G. Wang (2009) “Center-based sampling for population-based algorithms,” in: *Proceedings of IEEE Congress on Evolutionary Computation*, Trondheim, Norway, pp. 933-938.

H. R. Tizhoosh (2005) “Opposition-based learning: A new scheme for machine intelligence,” in: *Proceedings of International Conference on Computational Intelligence for Modelling, Control and Automation, and International Conference on Intelligent Agents, Web Technologies and Internet Commerce (CIMCA/IAWTIC-2005)*, pp. 695–701, 2005.

H.R. Tizhoosh (2006) “Reinforcement learning based on actions and opposite actions,” *Journal of Advanced Computational Intelligence and Intelligent Informatics*, Vol 10, No. 4, pp. 578-585.

M. Ventresca and H. R. Tizhoosh (2008) “A diversity maintaining population-based incremental learning algorithm,” *Information Sciences,* Vol. 178, No. 21, pp. 4038–4056.

M.Ventresca and H.R.Tizhoosh (2009), “Improving gradient-based learning algorithms for large scale feed forward networks,” in: *Proceedings of the International Joint Conference on Neural Networks,* 14–19 June, Atlanta, USA, pp. 3212–3219.

H. Wang, Z.J. Wu, Y. Liu, J. Wang, D. Z. Jiang and L. L. Chen, (2009) “Space transformation search: A new evolutionary technique,” in: *Proceedings of ACM/SIGEVO Summit on Genetic and Evolutionary Computation,* Shanghai China, pp. 537-544.

Q. Xu, L. Wang, B. M. He and N. Wang (2011) “Opposition-based differential evolution using the current optimum for function optimization,” *Journal of Applied Sciences,* Vol. 29, pp. 308-315.

Q. Xu, L. Wang, N. Wang and X. Hei, L. Zhao (2014a) “A review of opposition based learning from 2005 to 2012,” *Engineering Applications of Artificial Intelligence,* vol. 29, pp. 1–12.

Q. Xu, L. Guo N. Wang and Y. He (2014b), “COOBBO: a novel opposition-based soft computing algorithm for TSP problems,” *Algorithms*, Vol. 7, pp. 663-684.

Y. Zhou, J.-K. Hao and B. Duval (2017) “Opposition-based memetic search for the maximum diversity problem,” *IEEE Transactions on Evolutionary Computation*, DOI: 10.1109/TEVC.2017.2674800

## Appendix: Some Basic Diversification Generators for 0-1 Vectors

We indicate two basic types of diversification generators, one for problems that can be formulated in a natural manner as optimizing a function of zero-one variables, and the other for problems that can more appropriately be formulated as optimizing a permutation of elements. The generators described here are a subset of those identified in Glover (1997), and more advanced forms of these generators, along with additional types, can be found in Glover (2017).

The following approaches embody the precept that diversification is not the same as randomization, and hence differ from the randomized approaches for creating variation that are proposed in connection with a variety of evolutionary approaches. The goal of diversification is to produce solutions that differ from each other in significant ways, and that yield productive (or “interesting”) alternatives in the context of the problem considered. By contrast, the goal of randomization is to produce solutions that may differ from each other in any way (or to any degree) at all, as long as the differences are entirely “unsystematic”. From the present viewpoint, a reliance on variation that is strategically generated can offer advantages over a reliance on variation that is distinguished only by its unpredictability.

**Diversification Generators for Zero-One Vectors**

The first type of diversification generator takes a binary vector x as its seed solution, and generates a collection of solutions associated with an integer h = 1, 2,..., h*, where $h^* \le n - 1$. (Recommended is $h^* \le n/5$.)

We generate two types of solutions, **x'** and **x"** for each value of h, by the following rule:

*Type 1 Solution:* Let the first component $x_1'$ of **x'** be $1 - x_1$ and let $x_{1+kh}' = 1 - x_{1+kh}$ for k = 1, 2, 3,..., k*, where k* is the largest integer satisfying $k^* \le n/h$. Remaining components of **x'** equal 0.

To illustrate for **x** = (0,0,...,0): The values h = 1, 2 and 3 respectively yield **x'** = (1,1,...,1), (1,0,1,0,1 ...) and (1,0,0,1,0,0,1,0,0,1,....). This progression suggests the reason for preferring $h^* \le n/5$. As h becomes larger, the solutions **x'** for two adjacent values of h differ from each other proportionately less than when h is smaller. An option to exploit this is to allow h to increase by an increasing increment for larger values of h.

*Type 2 Solution:* Let **x"** be the complement of **x'**.

Again to illustrate for **x** = (0,0,...,0): the values h = 1, 2 and 3 respectively yield x"= (0,0,...,0), (0,1,0,1,....) and (0,1,1,0,1,1,0,...). Since x' duplicates x for h = 1, the value h = 1 can be skipped when generating **x"**.

The preceding design extends to generate additional solutions as follows. For values of $h \ge 3$ the solution vector is shifted so that the index 1 is instead represented as a variable index q, which can take the values 1, 2, 3, ..., h. Continuing the illustration for **x** = (0,0,...,0), suppose h = 3. Then, in addition to **x'** = (1,0,0,1,0,0,1,...), the method also generates the solutions given by **x'** = (0,1,0,0,1,0,0,1,...) and (0,0,1,0,0,1,0,0,1....), as q takes the values 2 and 3.

The following pseudo-code indicates how the resulting diversification generator can be structured, where the parameter MaxSolutions indicates the maximum number of solutions desired to be generated. Comments within the code appear in italics, enclosed within parentheses.

**First Diversification Generator for Zero-One Solutions.**

```
NumSolutions = 0
For h = 1 to h*
    Let q* = 1 if h < 3, and otherwise let q* = h
        (q* denotes the value such that q will range from 1 to q*. We set q* = 1 instead of q* = h
        for h < 3 because otherwise the solutions produced for the special case of h < 3 will
        duplicate other solutions or their complements.)
    For q = 1 to q*
        let k* = (n-q)/h <rounded down>
        For k = 1 to k*
            x_{q+kh}' = 1 – x_{q+kh}
        End k
```

If h > 1, generate **x"** as the complement of **x'**
*(**x'** and **x"** are the current output solutions.)*
NumSolutions = NumSolutions + 2 (or + 1 if h = 1)
If NumSolutions ≥ MaxSolutions, then stop generating solutions.
End q
End h

The number of solutions **x'** and **x"** produced by the preceding generator is approximately q*(q*+1). Thus if n = 50 and h* = n/5 = 10, the method will generate about 110 different output solutions, while if n = 100 and h* = n/5 = 20, the method will generate about 420 different output solutions.

Since the number of output solutions grows fairly rapidly as n increases, this number can be limited, while creating a relatively diverse subset of solutions, by allowing q to skip over various values between 1 and q*. The greater the number of values skipped, the less "similar" the successive solutions (for a given h) will be. Also, as previously noted, h itself can be incremented by a value that differs from 1.

*For added variation:*

If further variety is sought, the preceding approach can be augmented as follows. Let h = 3,4,..., h*, for h ≤ n - 2 (preferably h* ≤ n/3). Then for each value of h, generate the following solutions.

*Type 1A Solution:* Let $x_1' = 1 - x_1$ and $x_2' = 1 - x_2$. Thereafter, let $x_{1+kh}' = 1 - x_{1+kh}$ and let $x_{2+kh}' = 1 - x_{2+kh}$, for k = 1,2,...,k*, where k* is the largest integer such that 2 + kp ≤ n. All other components of **x'** are the same as in **x**.
*Type 2A Solution:* Create **x"** as the complement of **x'**, as before.

Related variants are evident. The index 1 can also be shifted (using a parameter q) in a manner similar to that indicated for solutions of type 1 and 2.

## A Sequential Diversification Generator

The concept of diversification invites a distinction between solutions that differ from a given solution (e.g., a seed solution) and those that differ from each other.[6] Our preceding comments refer chiefly to the second type of diversification, by their concern with creating a collection of solutions whose members exhibit certain contrasting features.

Diversification of the first type can be emphasized in the foregoing design by restricting attention to the complemented solutions denoted by **x"** when h becomes larger than 2. In general, diversification of the second type is supported by complementing larger numbers of variables in the seed solution. This type of diversification by itself is incomplete, and the relevance of diversification of the first type is important to heed in many situations.

[6] These distinctions have often been overlooked by the genetic algorithm community.

A sequential diversification generator for 0-1 vectors that follows the prescription to maximize the minimum distance from preceding vectors is embodied in the following procedure. We say that a solution **y** complements **x** over an index set J if $y_j = 1 - x_j$ for $j \in J$ and $y_j = x_j$ for $j \in N \backslash J$.

**Sequential (Max/Min) Diversification Generator**

1. Designate the seed solution **x** and its complement to be the first two solutions generated.
2. Partition the index set $N = \{1,\ldots,n\}$ for **x** into two sets N' and N" that, as nearly as possible, contain equal numbers of indexes. Create the two solutions **x'** and **x"** so that **x'** complements **x** over N' and **x"** complements **x** over N".
3. Define each subset of N that is created by the most recent partition of N to be a key subset. If no key subset of N contains more than 1 element, stop. Otherwise partition each key subset S of N into two sets S' and S" that contain, as nearly as possible, equal numbers of elements. (For the special case where S may contain only 1 element, designate one of S' and S" to be the same as S, and the other to be empty.) Overall, choose the designations S' and S" so that the number of partitions with |S'| > |S"| equals the number with |S"| > |S'|, as nearly as possible.
4. Let N' be the union of all subsets S' and let N" be the union of all subsets S". Create the complementary solutions **x'** and **x"** relative to N' and N" as in Step 2, and then return to Step 3. (The partition of each critical set into two parts in the preceding execution of Step 3 will cause the number of critical sets in the next execution of Step 3 to double.)

The foregoing process generates approximately $2(1 + \log n)$ solutions. If n is a power of 2, every solution produced maximizes the minimum Hamming distance from all previous solutions generated. (This maxmin distance, measured as the number of elements by which the solutions differ, is n/2 for every iteration after the first two solutions are generated in Step 1. Such a maxmin value is also approximately achieved when n is not a power of 2.)

In particular, starting with k = n, and updating k at the beginning of Step 3 by setting k: = k/2 (rounding fractional values upward), the number of elements in each key subset is either k or k-1. Thus, the method stops when k = 1. The balance between the numbers of sets S' and S" of different sizes can be achieved simply by alternating, each time a set S with an odd number of elements is encountered, in specifying the larger of the two members of the partition to be S' or S".

Useful variations result by partitioning N in different ways. Again, descriptions of these approaches may be found in Glover (1997).

## Diversification Generator for Permutation Problems

Although permutation problems can be formulated as 0-1 problems, they constitute a special class that preferably should be treated somewhat differently. Assume that a given trial permutation P used as a seed is represented by indexing its elements so they appear in consecutive order, to yield P = (1,2, ..., n). Define the subsequence P(h:s), where s is a positive integer between 1 and h, to be given by P(h:s) = (s, s+h, s+2h, ..., s+rh), where r is the largest nonnegative integer such that $s+rh \leq n$. Then define the permutation P(h), for $h \leq n$, to be P(h) = (P(h:h), P(h:h-1), ..., P(h:1)).

*Illustration:*

Suppose P is given by

P = (1, 2, 3, 4, 5, 6, 7, 8, 9, 10, 11, 12, 13, 14, 15, 16, 17, 18)

If we choose h = 5, then P(5:5) = (5,10,15), P(5:4) = (4,9,14), P(5:3) = (3,8,13,18), P(5:2) = (2,7,12,17), P(5:1) = (1,6,11,16), to give:

P(5) = (5, 10, 15, 4, 9, 14, 3, 8, 13, 18, 2, 7, 12, 17, 1, 6, 11, 16)

Similarly, if we choose h = 4 then P(4:4) = (4,8,12,16), P(4:3) = (3,7,11,15), P(4:2) = (2,6,10,14,18), P(4:1) = (1,5,9,13,17) to give:

P(4) = (4, 8, 12, 16, 3, 7, 11, 15, 2, 6, 10, 14, 18, 1, 5, 9, 13, 17)

In this illustration we have allowed h to take the two values closest to the square root of n. These values are interesting based on the fact that, when h equals the square root of n, the minimum relative separation of each element from each other element in the new permutation is maximum, compared to the relative separation of exactly 1 in the permutation P. In addition, other useful types of separation result, and become more pronounced for larger values of n.

In general, for the goal of generating a diverse set of permutations, preferable values for h range from 1 to n/2. We also generate the reverse of the preceding permutations, denoted by P*(h), which we consider to be more interesting than P(h). The preference of P*(h) to P(h) is greater for smaller values of h. For example, when h = 1, P(h) = P and P*(h) is the reverse of P. (Also, P(n) = P*(1).) In sum, we propose a Diversification Generator for permutation problems to be one that generates a subset of the collection P(h) and P*(h), for h = 1 to n/2 (excluding P(1) = P).